\documentclass{article}
\usepackage{ijcai22}

\usepackage{times}
\usepackage{soul}
\usepackage{url}
\usepackage[hidelinks]{hyperref}
\usepackage[utf8]{inputenc}
\usepackage[small]{caption}
\usepackage{graphicx}
\usepackage{amsmath}
\usepackage{amsthm}
\usepackage{amssymb} 
\usepackage{mathrsfs}
\usepackage{booktabs}
\usepackage{multirow}
\usepackage[ruled,vlined]{algorithm2e}
\usepackage{xcolor}

\newcommand{\name}[0]{BandMaxSAT\xspace}

\title{BandMaxSAT: A Local Search MaxSAT Solver with Multi-armed Bandit}

\author{
    Jiongzhi Zheng$^{1,2}$\and 
    Kun He\footnote{Corresponding author.}$^1$\and 
    Jianrong Zhou$^1$\and 
    Yan Jin$^1$\and 
    Chu-Min Li$^3$\and
    Felip Manya$^4$ 
    \affiliations
    $^1$School of Computer Science and Technology, Huazhong University of Science and Technology\\
    $^2$Institute of Artificial Intelligence, Huazhong University of Science and Technology, China\\
    $^3$MIS, University of Picardie Jules Verne, France\\
    $^4$Artificial Intelligence Research Institute (IIIA), CSIC, Bellaterra, Spain 
    \emails
    jzzheng@hust.edu.cn, brooklet60@hust.edu.cn
}

\begin{document}

\maketitle

\begin{abstract}
We address Partial MaxSAT (PMS) and Weighted PMS (WPMS), two practical generalizations of the MaxSAT problem, and propose a local search algorithm for these problems, called \name, that applies a multi-armed bandit model to guide the search direction. The bandit in our method is associated with all the soft clauses in the input (W)PMS instance. Each arm corresponds to a soft clause. The bandit model can help \name to select a good direction to escape from local optima by selecting a soft clause to be satisfied in the current step, that is, selecting an arm to be pulled. We further propose an initialization method for (W)PMS that prioritizes both unit and binary clauses when producing the initial solutions. Extensive experiments demonstrate that \name significantly outperforms the state-of-the-art (W)PMS local search algorithm SATLike3.0. Specifically, the number of instances in which \name obtains better results is about twice that obtained by SATLike3.0. Moreover, we combine \name with the complete solver TT-Open-WBO-Inc. The resulting solver \name-c also outperforms some of the best state-of-the-art complete (W)PMS solvers, including SATLike-c, Loandra and TT-Open-WBO-Inc.
\end{abstract}


\section{Introduction}

As an optimization extension of the famous Boolean Satisfiability (SAT) decision problem, the Maximum Satisfiability (MaxSAT) problem aims at finding a complete assignment of the Boolean variables to satisfy as many clauses as possible in a given propositional formula in Conjunctive Normal Form (CNF)~\cite{li2021maxsat}. Partial MaxSAT (PMS) is a variant of MaxSAT where the clauses are divided into hard and soft. PMS aims at maximizing the number of satisfied soft clauses with the constraint that all the hard clauses must be satisfied. Associating a positive weight to each soft clause in PMS results in Weighted PMS (WPMS), whose goal is to maximize the total weight of satisfied soft clauses with the same constraint of PMS that all the hard clauses must be satisfied. Both PMS and WPMS, denoted as (W)PMS, have many practical applications such as planning~\cite{Bonet2019}, combinatorial testing~\cite{AMOST22}, group testing~\cite{Ciampiconi2020}, and timetabling~\cite{Demirovic2017}.

In this paper, we focus on the local search approach, which is a well-studied category of incomplete (W)PMS algorithms and exhibits promising performance on random and crafted (W)PMS instances. Recent well-performing (W)PMS local search algorithms, such as Dist~\cite{Cai2014}, CCEHC~\cite{Luo2017}, SATLike~\cite{Lei2018} and SATLike3.0~\cite{Cai2020}, all start from an initial complete assignment and then flip the Boolean value of a selected variable per step to find better solutions. These local search algorithms follow similar procedures to escape from local optima. Note that a local optimum indicates that flipping any single variable cannot improve the current solution. 

When falling into an infeasible local optimum (i.e., there are falsified hard clauses), these algorithms first randomly select a falsified hard clause and then satisfy it by flipping one of its variables. The random strategy for selecting the falsified hard clause is reasonable, since all the hard clauses should be satisfied. However, when falling into a feasible local optimum (i.e., there are no falsified hard clauses), these algorithms still use the random strategy to determine the soft clause to be satisfied in the current step, which may not be a good strategy for the following reasons: 1) different from hard clauses, not all the soft clauses should be satisfied. 2) the high degree of randomness may lead to a small probability for these algorithms to find a good search direction (satisfying a falsified soft clause corresponds to a search direction).

To handle the above issues, we propose a multi-armed bandit (MAB) local search algorithm, called \name, for (W)PMS. MAB is a basic model in the field of reinforcement learning~\cite{Slivkins2019,Lattimore2020}. In an MAB reinforcement learning model, the agent needs to select to pull an arm (i.e., perform an action) at each decision step (i.e., state), which leads to some rewards. The agent uses the rewards to evaluate the benefit of pulling each arm and uses the evaluation values to decide the arm to be pulled in each step. In summary, the MAB can be used to help a program learn to select an appropriate item from multiple candidates. Therefore, we propose to apply an MAB to help the (W)PMS local search algorithm learn to select an appropriate soft clause (i.e., a high-quality search direction) to be satisfied, whenever the search falls into a feasible local optimum. Specifically, each arm in the bandit of \name corresponds to a soft clause in the input (W)PMS instance. Pulling an arm implies selecting the corresponding clause to be satisfied in the current step.


There are related studies that apply MAB to MaxSAT. For example, Goffinet and Ramanujan~[\citeyear{Goffinet2016}] proposed an algorithm for MaxSAT, based on Monte-Carlo tree search, where a two-armed bandit is associated with each variable (node in the search tree) to decide the branching direction, i.e., which Boolean value  assign to the variable. Their approach needs a local search solver to evaluate the quality of the branching nodes, thus the performance relies on the local search solver. Lassouaoui et al.~[\citeyear{Lassouaoui2019}] proposed to use an MAB model to select the low-level heuristics in a hyper-heuristic framework for MaxSAT. Pulling an arm in their model implies selecting a corresponding low-level heuristic to optimize the current solution. They didn't compare with the state-of-the-art (in)complete MaxSAT algorithms, but only compared with other hyper-heuristic methods. Our work proposes a novel MAB model for (W)PMS that significantly improves (W)PMS state-of-the-art local search methods. To our knowledge, this is the first time 
that an MAB model is associated with the clauses in a (W)PMS local search solver.

Moreover, inspired by the studies for SAT and MaxSAT that prioritize both unit and binary clauses (i.e., clauses with exactly one and two literals, respectively) over other clauses~\cite{Chao1990,Chvatal1992,Li2006}, we propose a novel decimation approach that prefers to satisfy both unit and binary clauses, denoted as hybrid decimation (HyDeci), for generating the initial assignment in \name. The decimation method is a category of incomplete approaches that proceeds by assigning the Boolean value of some (usually one) variables sequentially and simplifies the formula accordingly~\cite{Cai2017}. Decimation approaches that focus on unit clauses have been used in MaxSAT~\cite{Cai2017,Cai2020}. However, it is the first time, to our knowledge, that a decimation method concentrating on both unit and binary clauses is used in MaxSAT. The experimental results demonstrate that considering both unit and binary clauses is better than only considering unit clauses.

To evaluate the performance of the proposed \name algorithm, we compare \name with the state-of-the-art (W)PMS local search algorithm  SATLike3.0~\cite{Cai2020}. The experiments show that \name significantly outperforms SATLike3.0 on both PMS and WPMS. Moreover, as one of the state-of-the-art (W)PMS solvers, SATLike-c~\cite{Lei2021} combines SATLike3.0 with an effective complete solver, TT-Open-WBO-Inc~\cite{Nadel2019}, and won three categories among the total four (PMS and WPMS categories, each associated with two time limits) of the incomplete track in the latest MaxSAT Evaluation (MSE2021). By combining \name with TT-Open-WBO-Inc, the resulting solver \name-c also outperforms some of the best state-of-the-art (W)PMS complete solvers, including SATLike-c, Loandra~\cite{Berg2019}, and TT-Open-WBO-Inc.

The main contributions of this work are as follows:

\begin{itemize}
\item We propose a  multi-armed bandit (MAB) model that fits well with the MaxSAT local search algorithms, and an effective local search solver for (W)PMS, called \name, that applies the proposed bandit model to guide the search direction.
\item We demonstrate that there is a great potential to use MAB in MaxSAT solving. Our proposed MAB model is general and could be applied to improve other MaxSAT local search algorithms.
\item We propose a novel decimation method for (W)PMS, denoted as HyDeci, that prefers to satisfy both unit and binary clauses. HyDeci provides high-quality initial assignments for \name, and could be applied to improve other MaxSAT local search algorithms.
\item Extensive experiments show that \name significantly outperforms the state-of-the-art (W)PMS local search algorithm SATLike3.0. Moreover, by combining \name with the complete solver TT-Open-WBO-Inc, the resulting solver \name-c also outperforms the state-of-the-art (W)PMS complete solvers.
\end{itemize}

\section{Preliminaries}
Given a set of Boolean variables $\{x_1,...,x_n\}$, a literal is either a variable itself $x_i$ or its negation $\lnot x_i$; a clause is a disjunction of literals, i.e., $c_j = l_{j1} \lor ... \lor l_{j n_j}$, where $n_j$ is the number of literals in clause $c_j$. A  Conjunctive Normal Form (CNF) formula $\mathcal{F}$ is a conjunction of clauses, i.e., $\mathcal{F} = c_1 \land ... \land c_m$. A complete assignment $A$ represents a mapping that maps each variable to a value of 1 (true) or 0 (false). A literal $x_i$ (resp. $\lnot x_i$) is true if the current assignment maps $x_i$ to 1 (resp. 0). A clause is satisfied by the current assignment if there is at least one true literal in the clause. 

Given a CNF formula $\mathcal{F}$,
MaxSAT 
aims at finding an assignment that satisfies as many clauses in $\mathcal{F}$ as possible. Given a CNF formula $\mathcal{F}$ whose clauses are divided into hard and soft, PMS is a variant of MaxSAT that aims at finding an assignment that satisfies all the hard clauses and maximize the number of satisfied soft clauses in $\mathcal{F}$, and WPMS is a generalization of PMS where each soft clause is associated with a positive weight. The goal of WPMS is to find an assignment that satisfies all the hard clauses and maximizes the total weight of satisfied clauses in $\mathcal{F}$. In the local search algorithms for (Max)SAT, the flipping operator for a variable is an operator that changes its Boolean value.


Given a (W)PMS instance $\mathcal{F}$, a complete assignment $A$ is feasible if it satisfies all the hard clauses in $\mathcal{F}$. The cost of $A$, denoted as $cost(A)$, is set to $+\infty$ for convenience if $A$ is infeasible. Otherwise, $cost(A)$ is equal to the number of falsified soft clauses for PMS, and equal to the total weight of falsified soft clauses for WPMS.

In addition, the effective clause weighting technique is widely used in recent well-performing (W)PMS local search algorithms~\cite{Cai2014,Luo2017,Cai2020}. Algorithms with this technique associate dynamic weights (independent of the original soft clause weights in WPMS instances) to clauses and use the dynamic weights to guide the search direction. \name also applies the clause weighting technique, and maintains dynamic weights to both hard clauses and soft clauses with the clause weighting strategy used in SATLike3.0~\cite{Cai2020}. 

Given a (W)PMS instance $\mathcal{F}$, the current assignment $A$, and the dynamic clause weights, the commonly used scoring function for a variable $x$, denoted as $score(x)$, is defined as the increment or reduction of the total dynamic weight of satisfied clauses caused by flipping $x$ in $A$. Moreover, a local optimum for (W)PMS indicates that there are no variables with positive \textit{score}. A local optimum is feasible if there are no falsified hard clauses, otherwise it is infeasible.

\section{Methodology}
The proposed local search algorithm \name consists of the proposed hybrid decimation (HyDeci) initialization process and the search process. During the local search process, we use a multi-armed bandit that is associated with the soft clauses to help \name learn to select good directions to escape from feasible local optima. This section first introduces the HyDeci method and the bandit model used in \name, and then the main process of \name.

\subsection{Hybrid Decimation}
HyDeci is an effective decimation method that prefers to satisfy both unit and binary clauses. Since the clauses with shorter lengths are easier to be falsified, preferring to satisfy shorter clauses can reduce the number of falsified clauses, which results in high-quality initial assignments. The procedure of HyDeci is shown in Algorithm \ref{alg:HyDeci}. We use SIMPLIFY to refer to the process of simplifying the formula after assigning a value to a variable.

\begin{algorithm}[t]
\caption{HyDeci($\mathcal{F}$)}
\label{alg:HyDeci}
\LinesNumbered 
\KwIn{A (W)PMS instance $\mathcal{F}$}
\KwOut{A complete assignment $A$ of variables in $\mathcal{F}$}
\While{$\exists$ unassigned variables}{
\If{$\exists$ hard unit clauses}{
$c :=$ a random hard unit clause\;
satisfy $c$ and SIMPLIFY\;
}
\ElseIf{$\exists$ soft unit clauses}{
$c :=$ a random soft unit clause\;
satisfy $c$ and SIMPLIFY\;
}
\ElseIf{$\exists$ hard binary clauses}{
$c :=$ a random hard binary clause\;
$l :=$ a greedily selected unassigned literal in $c$\;
satisfy $l$ and SIMPLIFY\;
}
\ElseIf{$\exists$ soft binary clauses}{
$c :=$ a random soft binary clause\;
$l :=$ a greedily selected unassigned literal in $c$\;
satisfy $l$ and SIMPLIFY\;
}
\Else{$v :=$ a random unassigned variable\;
assign $v$ a random value and SIMPLIFY\;
}
}
\textbf{return} the resulting complete assignment $A$\;
\end{algorithm}

HyDeci generates the initial complete assignment iteratively. In each iteration, HyDeci assigns the value of exactly one variable. When there are unit clauses, HyDeci samples a random unit clause (hard clauses take precedence) and then satisfies it. When there is no unit clause but there are binary clauses, HyDeci first samples a random binary clause $c$ (hard clauses take precedence), and then selects one of the two unassigned literals in $c$ and satisfies it according to a greedy strategy, that is, preferring to satisfy the literal whose satisfaction leads to more satisfied soft clauses (or to a larger total weight of satisfied soft clauses). When there are no unit and binary clauses, HyDeci randomly selects an unassigned variable and randomly assigns a Boolean value to it. 

The main improvement of the proposed HyDeci algorithm over the previous decimation approaches~\cite{Cai2017,Cai2020} is that HyDeci not only concentrates on unit clauses but also on binary clauses. The experimental results demonstrate that considering both unit and binary clauses is better than only considering unit clauses to generate high-quality initial assignments.

\subsection{Multi-armed Bandit Model for (W)PMS}
We propose a multi-armed bandit model for (W)PMS to help \name learn to select the appropriate soft clause to be satisfied when falling into a feasible local optimum. Each arm of the bandit model corresponds to a soft clause. Pulling an arm implies selecting the corresponding soft clause to be satisfied in the current step. The bandit model maintains an estimated value $V(i)$ and a selected time $t(i)$ for each arm (i.e., soft clause) $i$. We initialize $V(i) = 1$ and $t(i) = 0$ for each arm $i$. The larger the estimated value of an arm, the more benefits of pulling the arm, i.e., satisfying the soft clause corresponding to the arm may yield better solutions. 

The rest of this subsection first introduces the method of selecting an arm to be pulled and then the method of updating the estimated values.

\subsubsection{Arm Selection Strategy}
\name uses the Upper Confidence Bound method~\cite{Hu2019} to trade-off between exploration and exploitation and selects the arm to be pulled. Specifically, the upper confidence bound $U_i$ on the estimated value $V_i$ of arm $i$ is calculated with the following equation:

\begin{equation}
U_i = V_i + \lambda \cdot \sqrt{\frac{ln(N)}{t(i) + 1}},
\label{eq:UCB}
\end{equation}
where $N$ indicates the number of times fallen into a feasible local optimum and $\lambda$ is the exploration bias parameter. 

The procedure of selecting the arm is shown in Algorithm \ref{alg:PickArm}. Since our bandit contains a large number of arms (equal to the number of soft clauses), selecting the best among all the arms is inefficient. Therefore, \name first applies (line 3) the sampling strategy to randomly sample $ArmNum$ (20 by default) candidate arms and then selects the arm with the highest upper confidence bound among the candidates (lines 4-5). Similar sampling strategies have been used in multi-armed bandit problems~\cite{Ou2019} and some combinatorial optimization problems~\cite{Cai2015}. Note that the bandit aims at selecting a soft clause to be satisfied in the current step. Thus, the arms corresponding to the soft clauses that are satisfied by the current assignment will not be considered as candidates. The experimental results show that the sampling strategy in our bandit model can significantly improve the algorithm's performance.

\begin{algorithm}[t]
\caption{PickArm($ArmNum,N,\lambda$)}
\label{alg:PickArm}
\LinesNumbered 
\KwIn{Number of sampled arms $ArmNum$, number of times to fall into a local optimum $N$, exploration bias parameter $\lambda$}
\KwOut{The arm selected to be pulled $c$}
initialize $U^* := -\infty$\;
\For{$i := 1$ to $ArmNum$}{
$j :=$ a random falsified soft clause\;
calculate $U_j$ according to Eq. \ref{eq:UCB}\;
\lIf{$U_j > U^*$}{$U^* := U_j$, $c := j$}
}
\textbf{return} $c$\;
\end{algorithm}

\subsubsection{Estimated Value Updating Strategy}
Since \name pulls an arm of the bandit when it falls into a feasible local optimum, we apply the \textit{cost} values of the feasible local optimal solutions before and after pulling an arm to calculate the reward of the action (i.e., pulling the arm). Suppose $A'$ and $A$ are the last and current feasible local optimal solutions respectively, and $c$ is the last pulled arm, a simple reward for pulling $c$ can be set to $r = cost(A') - cost(A)$. However, reducing the \textit{cost} value from 20 to 10 is much harder and more meaningful than reducing it from 1000 to 990. Thus, the rewards of these two cases should not be the same. 
To address this issue, 
we define the reward as follows:
\begin{equation}
r(A,A',A^*) = \frac{cost(A') - cost(A)}{cost(A') - cost(A^*) + 1},
\label{eq:reward}
\end{equation}
where $A^*$ is the best solution found so far. 
Suppose in Eq. \ref{eq:reward} $cost(A') - cost(A)$ is constant, then the closer $cost(A')$ and $cost(A^*)$, the more rewards the action of pulling the last arm can yield, which is reasonable and intuitive.

Moreover, since the arms (i.e., soft clauses) are connected by the variables, we assume that the arms in our bandit model are not independent of each other. We also believe that the improvement (or deterioration) of $A$ over $A'$ may not only be due to the last action, but also due to earlier actions. 
Hence, 
we apply the delayed reward method~\cite{Arya2019} to update the estimated value of the last $d$ (20 by default) pulled arms once a reward is obtained. Specifically, suppose that $A'$ and $A$ are the last and current feasible local optimal solutions respectively, $A^*$ is the best solution found so far, and $\{a_1,...,a_d\}$ is the set of the latest $d$ pulled arms ($a_d$ is the most recent one). Then, the estimated values of the $d$ arms are updated as follows:
\begin{equation}
V_{a_i} = V_{a_i} + \gamma^{d-i} \cdot r(A,A',A^*), i \in \{1,...,d\},
\label{eq:updateV}
\end{equation}
where $\gamma$ is the reward discount factor and  $r(A,A',A^*)$ is calculated with Eq.~\ref{eq:reward}. 

\begin{algorithm}[t]
\caption{\name}
\label{alg:PPMAB}
\LinesNumbered 
\KwIn{A (W)PMS instance $\mathcal{F}$, cut-off time \textit{cutoff}, BMS parameter $k$, reward delay steps $d$, reward discount factor $\gamma$, number of sampled arms $ArmNum$, exploration bias parameter $\lambda$}
\KwOut{A feasible assignment $A$ of $\mathcal{F}$, or \textit{no feasible assignment found}}
$A :=$ HyDeci($\mathcal{F}$)\;
$A^* := A$, $cost(A') := +\infty$, $N := 0$\;
\While{\textit{running time} $<$ \textit{cutoff}}{
\If{$A$ is feasible $\&$ \textit{cost}($A$) $<$ \textit{cost}($A^*$)}{
$A^* := A$\;
}
\eIf{$D := \{x|score(x)>0\} \neq \emptyset$}{
$v :=$ a variable in $D$ picked by BMS($k$)\;
}
{update\_clause\_weights()\;
\eIf{$\exists$ \textit{falsified hard clauses}}
{$c :=$ a random falsified hard clause\;}
{
update\_estimated\_value($A,A',A^*,d,\gamma$)\;
$N := N + 1$, $A' := A$\;
$c :=$ PickArm($ArmNum,N,\lambda$)\;
$t(c) := t(c) + 1$\;
}
$v :=$ the variable with the highest \textit{score} in $c$\;
}
$A := A$ with $v$ flipped\;
}
\lIf{$A^*$ is feasible}{\textbf{return} $A^*$}
\lElse{\textbf{return} \textit{no feasible assignment found}}
\end{algorithm}

\subsection{Main Process of BandMaxSAT}
Finally, we introduce the main process of \name, which is shown in Algorithm \ref{alg:PPMAB}. \name first uses HyDeci to generate an initial assignment, and then repeatedly selects a variable and flips it until the cut-off time is reached.

When local optima are not reached, \name selects a variable to be flipped using the Best from Multiple Selections (BMS) strategy~\cite{Cai2015}. BMS chooses $k$ random variables (with replacement) and returns one with the highest \textit{score} (lines 6-7). When falling into a local optimum, \name first updates the dynamic clause weights (by the update\_clause\_weight() function in line 9) according to the clause weighting scheme in SATLike3.0~\cite{Cai2020}, and then selects the clause to be satisfied in the current step.

If the local optimum is infeasible, \name randomly selects a falsified hard clause as the clause to be satisfied in the current step (lines 10-11). If the local optimum is feasible (lines 12-15), \name first updates the estimated values of the latest $d$ pulled arms according to Eq. \ref{eq:updateV}, and then uses the PickArm() function (Algorithm \ref{alg:PickArm}) to select the soft clause to be satisfied in the current step. After determining the clause $c$ to be satisfied, \name flips the variable with the highest \textit{score} in $c$ in the current step (line 16).


\section{Experimental Results}
We compare BandMaxSAT with the state-of-the-art (W)PMS local search algorithm, SATLike3.0~\cite{Cai2020}, as well as some of the best state-of-the-art complete solvers SATLike-c~\cite{Lei2021}, Loandra~\cite{Berg2019} and TT-Open-WBO-Inc~\cite{Nadel2019}. The experimental results demonstrate the excellent performance of our proposed BandMaxSAT algorithm, that significantly improves the best (W)PMS local search solver. 
The results also show the effectiveness of each component in BandMaxSAT, including the HyDeci initialization method, the sampling strategy and the delayed reward method.

\subsection{Experimental Setup}
BandMaxSAT is implemented in C++ and compiled by g++. Our experiments were performed on a server using an Intel® Xeon® E5-2650 v3 2.30 GHz 10-core CPU and 256 GB RAM, running Ubuntu 16.04 Linux operation system. We tested the algorithms on all the (W)PMS instances from the incomplete track of last four MaxSAT Evaluations (MSE), i.e., MSE2018-MSE2021. Note that we denote the benchmark that contains all the PMS/WPMS instances from the incomplete track of MSE2021 as PMS\_2021/WPMS\_2021, and so on. Each instance is calculated once by each algorithm with a time limit of 300 seconds, which is consistent with the settings in MSEs and \cite{Cai2020}. The best results in the tables appear in bold. Moreover, we use BandMS as a short name of BandMaxSAT in the tables.

The parameters in BandMaxSAT include the BMS parameter $k$, the reward delay steps $d$, the reward discount factor $\gamma$, the number of sampled arms $ArmNum$, and the exploration bias parameter $\lambda$. We use all the (W)PMS instances from the incomplete track of MSE2017 as the training set to tune these parameters. The parameter domains of these parameters are as follows: $[10,50]$ for $k$, $[1,50]$ for $d$, $[0.5,1]$ for $\gamma$, $[10,50]$ for $ArmNum$, and $[0.1,10]$ for $\lambda$. Finally, the default settings of these parameters are as follows: $k = 15$, $d = 20$, $\gamma = 0.9$, $ArmNum = 20$, $\lambda = 1$. The code of BandMaxSAT is available at https://github.com/JHL-HUST/BandMaxSAT/.

\subsection{Comparison of BandMaxSAT and SATLike3.0}
We first compare BandMaxSAT with the state-of-the-art (W)PMS local search algorithm, SATLike3.0~\cite{Cai2020}, on all the tested instances. The results are shown in Table \ref{table-BandMS-SL}. Column \textit{\#inst.} indicates the number of instances in each benchmark. Column \textit{\#win.} indicates the number of instances in which the algorithm yields the best solution among all the algorithms in the table. Column \textit{time} represents the average running time to yield the \textit{\#win.} instances.

As shown by the results in Table \ref{table-BandMS-SL}, BandMaxSAT (BandMS) significantly outperforms SATLike3.0 for both PMS and WPMS. Specifically, the \textit{\#win.} instances of BandMaxSAT is 62-131\% greater than those of SATLike3.0 for WPMS, and 43-103\% greater than those of SATLike3.0 for PMS, indicating a significant improvement. 


\begin{table}[t]
\footnotesize
\centering
\scalebox{1.0}{\begin{tabular}{lrrrrrr} \toprule
\multicolumn{1}{l}{\multirow{2}{*}{Benchmark}} & \multirow{2}{*}{\#inst.} & \multicolumn{2}{r}{BandMS}                                    &  & \multicolumn{2}{r}{SATLike3.0}                      \\ \cline{3-4} \cline{6-7} 
\multicolumn{1}{c}{}                           &                          & \hspace{-0.5em}\#win.                          & time                       &  & \#win.                 & time                       \\ \hline
WPMS\_2018                                     & 172                      & \textbf{118}                    & 114.76                    &  & 51                     & 73.68                     \\
WPMS\_2019                                     & 297                      & \textbf{210}                    & 108.47                    &  & 103                    & 75.50                     \\
WPMS\_2020                                     & 253                      & \textbf{170}                    & 137.45                    &  & 88                     & 81.81                     \\
WPMS\_2021                                     & 151                      & \textbf{89}                     & 145.87                    &  & 55                     & 89.70                     \\
PMS\_2018                                      & 153                      & \textbf{110}                    & 99.15                     &  & 54                     & 80.39                     \\
PMS\_2019                                      & 299                      & \textbf{204}                    & 67.93                     &  & 143                    & 53.67                     \\
PMS\_2020                                      & 262                      & \textbf{174}                    & 73.91                     &  & 119                    & 63.18                     \\
PMS\_2021                                      & 155                      & \textbf{112}                    & 68.66                     &  & 60                     & 51.77          \\ \bottomrule          
\end{tabular}}
\vspace{-0.4em}
\caption{Comparison of BandMS and SATLike3.0.}
\label{table-BandMS-SL}
\vspace{-0.4em}
\end{table}

\begin{table}[t]
\footnotesize
\centering
\scalebox{1.0}{\begin{tabular}{lrrrr} \toprule
Benchmark                 & \hspace{-0.5em}BandMS-c         & \hspace{-0.5em}SATLike-c       & Loandra & TT-OWI          \\ \hline
WPMS\_2018                & \textbf{0.9041} & 0.8901          & 0.8820  & 0.9026          \\
WPMS\_2019                & 0.8974          & 0.8819          & 0.8474  & \textbf{0.9022} \\
WPMS\_2020                & \textbf{0.8707} & 0.8613          & 0.8043  & 0.8655          \\
WPMS\_2021                & \textbf{0.7844} & 0.7809          & 0.7833  & 0.7729          \\
PMS\_2018                 & \textbf{0.8537} & 0.8440          & 0.7811  & 0.8412          \\
PMS\_2019                 & \textbf{0.8793} & 0.8745          & 0.7818  & 0.8713          \\
PMS\_2020                 & \textbf{0.8619} & 0.8511          & 0.8216  & 0.8586          \\
PMS\_2021                 & \textbf{0.8592} & 0.8538          & 0.7714  & 0.8436       \\ \bottomrule 
\end{tabular}}
\vspace{-0.4em}
\caption{Comparison of BandMS-c and the state-of-the-art complete (W)PMS solvers, SATLike-c, Loandra, TT-Open-WBO-Inc. The results are expressed by the scoring function used in MSE2021.}
\label{table-BandMS-c}
\vspace{-0.8em}
\end{table}

\begin{table}[t]
\footnotesize
\centering
\begin{tabular}{lrrrrrrrrr} \toprule
\multicolumn{1}{l}{\multirow{2}{*}{Benchmark}} & \multirow{2}{*}{\hspace{-1.5em}\#inst.} & \multicolumn{2}{r}{\hspace{-1em}BandMS} & \hspace{-0.5em} & \multicolumn{2}{r}{\hspace{-1em}Sample-1} & \hspace{-0.5em} & \multicolumn{2}{r}{\hspace{-1em}Sample-all} \\ \cline{3-4} \cline{6-7} \cline{9-10} 
\multicolumn{1}{c}{}                           &                          & \hspace{-1.1em}\#win.         & \hspace{-1em}time     & \hspace{-1em} & \hspace{-1em}\#win.          & \hspace{-1em}time             & \hspace{-1em} & \hspace{-1em}\#win.            & \hspace{-1em}time            \\ \hline
WPMS\_2018                                     & \hspace{-1em}172                      & \hspace{-1.1em}\textbf{100}   & \hspace{-1em}102.04  & \hspace{-1em} & \hspace{-1em}62              & \hspace{-1em}66.47           & \hspace{-1em} & \hspace{-1em}65                & \hspace{-1em}74.15           \\
WPMS\_2019                                     & \hspace{-1em}297                      & \hspace{-1.1em}\textbf{188}   & \hspace{-1em}100.35  & \hspace{-1em} & \hspace{-1em}117             & \hspace{-1em}73.21           & \hspace{-1em} & \hspace{-1em}118               & \hspace{-1em}78.91           \\
WPMS\_2020                                     & \hspace{-1em}253                      & \hspace{-1.1em}\textbf{160}   & \hspace{-1em}127.43  & \hspace{-1em} & \hspace{-1em}96              & \hspace{-1em}81.86           & \hspace{-1em} & \hspace{-1em}100               & \hspace{-1em}94.54           \\
WPMS\_2021                                     & \hspace{-1em}151                      & \hspace{-1.1em}\textbf{80}    & \hspace{-1em}134.94  & \hspace{-1em} & \hspace{-1em}53              & \hspace{-1em}114.89          & \hspace{-1em} & \hspace{-1em}46                & \hspace{-1em}90.91           \\
PMS\_2018                                      & \hspace{-1em}153                      & \hspace{-1.1em}\textbf{102}   & \hspace{-1em}87.75   & \hspace{-1em} & \hspace{-1em}74              & \hspace{-1em}59.45           & \hspace{-1em} & \hspace{-1em}67                & \hspace{-1em}65.78           \\
PMS\_2019                                      & \hspace{-1em}299                      & \hspace{-1.1em}\textbf{191}   & \hspace{-1em}65.35   & \hspace{-1em} & \hspace{-1em}164             & \hspace{-1em}53.26           & \hspace{-1em} & \hspace{-1em}130               & \hspace{-1em}59.26           \\
PMS\_2020                                      & \hspace{-1em}262                      & \hspace{-1.1em}\textbf{168}   & \hspace{-1em}67.38   & \hspace{-1em} & \hspace{-1em}139             & \hspace{-1em}54.28           & \hspace{-1em} & \hspace{-1em}111               & \hspace{-1em}66.21           \\
PMS\_2021                                      & \hspace{-1em}155                      & \hspace{-1.1em}\textbf{109}   & \hspace{-1em}65.61   & \hspace{-1em} & \hspace{-1em}83              & \hspace{-1em}36.24           & \hspace{-1em} & \hspace{-1em}72                & \hspace{-1em}49.41          \\ \bottomrule
\end{tabular}
\vspace{-0.4em}
\caption{Comparison with variants Sample-1 and Sample-all.}
\label{table-BandMS-sample}
\vspace{-0.7em}
\end{table}

\begin{table}[t]
\footnotesize
\centering
\scalebox{1.0}{\begin{tabular}{lrrrrrr} \toprule
\multicolumn{1}{l}{\multirow{2}{*}{Benchmark}} & \multirow{2}{*}{\#inst.} & \multicolumn{2}{r}{BandMS} &  & \multicolumn{2}{r}{BandMS$_{\text{no-delay}}$} \\ \cline{3-4} \cline{6-7} 
\multicolumn{1}{c}{}                           &                          & \#win.         & time     &  & \#win.          & time            \\ \hline
WPMS\_2018                                     & 172                      & \textbf{117}   & 108.96  &  & 85              & 64.60          \\
WPMS\_2019                                     & 297                      & \textbf{187}   & 106.78  &  & 165             & 87.30          \\
WPMS\_2020                                     & 253                      & \textbf{161}   & 130.91  &  & 133             & 91.50          \\
WPMS\_2021                                     & 151                      & \textbf{89}    & 140.16  &  & 64              & 112.62         \\
PMS\_2018                                      & 153                      & \textbf{113}   & 91.99   &  & 97              & 57.00          \\
PMS\_2019                                      & 299                      & \textbf{202}   & 66.14   &  & 191             & 58.40          \\
PMS\_2020                                      & 262                      & \textbf{181}   & 71.78   &  & 158             & 43.55          \\
PMS\_2021                                      & 155                      & \textbf{108}   & 60.98    &  & 97              & 48.91          \\ \bottomrule
\end{tabular}}
\vspace{-0.4em}
\caption{Comparison with variant BandMS$_{\text{no-binary}}$.}
\label{table-BandMS-delay}
\vspace{-0.8em}
\end{table}

\begin{table}[t]
\footnotesize
\centering
\scalebox{1.0}{\begin{tabular}{lrrrrrr} \toprule
\multicolumn{1}{l}{\multirow{2}{*}{Benchmark}} & \multirow{2}{*}{\#inst.} & \multicolumn{2}{r}{BandMS}               &  & \multicolumn{2}{r}{BandMS$_{\text{no-binary}}$}      \\ \cline{3-4} \cline{6-7} 
\multicolumn{1}{c}{}                           &                          & \#win.       & time &  & \#win.       & time \\ \hline
WPMS\_2018                                     & 172                      & \textbf{105} & 110.11                  &  & 86           & 94.17                   \\
WPMS\_2019                                     & 297                      & \textbf{190} & 104.24                  &  & 165          & 93.96                   \\
WPMS\_2020                                     & 253                      & \textbf{161} & 129.54                  &  & 135          & 103.39                  \\
WPMS\_2021                                     & 151                      & \textbf{89}  & 146.36                  &  & 76           & 119.98                  \\
PMS\_2018                                      & 153                      & \textbf{100} & 107.80                  &  & 98           & 79.07                   \\
PMS\_2019                                      & 299                      & \textbf{203} & 76.28                   &  & 201          & 63.91                   \\
PMS\_2020                                      & 262                      & \textbf{178} & 80.20                   &  & 176          & 63.13                   \\
PMS\_2021                                      & 155                      & \textbf{102} & 72.00                   &  & 92           & 42.79                  \\ \bottomrule
\end{tabular}}
\vspace{-0.4em}
\caption{Comparison with variant BandMS$_{\text{no-binary}}$.}
\label{table-BandMS-binary}
\vspace{-0.8em}
\end{table}
\begin{table}[t]
\footnotesize
\centering
\scalebox{1.0}{\begin{tabular}{lrrrrrr} \toprule
\multicolumn{1}{l}{\multirow{2}{*}{Benchmark}} & \hspace{-0.5em}\multirow{2}{*}{\#inst.} & \multicolumn{2}{r}{BandMS$_{\text{fast}}$}            &  & \multicolumn{2}{r}{BandMS$_{\text{no-binary-fast}}$}             \\ \cline{3-4} \cline{6-7} 
\multicolumn{1}{c}{}                           &                          & \#win.       & time                       &  & \#win.                 & time                       \\ \hline
WPMS\_2018                                     & 172                      & \textbf{91}  & 9.70                      &  & 89                     & 3.15                      \\
WPMS\_2019                                     & 297                      & 162          & 10.80                     &  & \textbf{165}           & 7.31                      \\
WPMS\_2020                                     & 253                      & \textbf{150} & 13.18                     &  & 136                    & 11.36                     \\
WPMS\_2021                                     & 151                      & \textbf{90}  & 18.49 &  & 69 & 15.62 \\
PMS\_2018                                      & 153                      & \textbf{94}  & 6.15                      &  & 90                     & 4.64                      \\
PMS\_2019                                      & 299                      & \textbf{188} & 5.06                      &  & 168                    & 4.72                      \\
PMS\_2020                                      & 262                      & \textbf{166} & 4.62                      &  & 147                    & 3.75                      \\
PMS\_2021                                      & 155                      & \textbf{89}  & 7.17                      &  & 79                     & 5.13         \\ \bottomrule            
\end{tabular}}
\vspace{-0.4em}
\caption{Comparison with  BandMS$_{\text{fast}}$ and~BandMS$_{\text{no-binary-fast}}$.}
\label{table-BandMS-fast}
\vspace{-0.8em}
\end{table}

\subsection{Comparison with Complete Solvers}
We then combine our BandMaxSAT local search algorithm with the complete solver TT-Open-WBO-Inc (TT-OWI) as SATLike-c does (which combines SATLike3.0 with TT-OWI), and denote the resulting solver as BandMaxSAT-c (BandMS-c). We further compare BandMaxSAT-c with some of the best state-of-the-art complete (W)PMS solvers that showed excellent performance in recent MSEs, including SATLike-c, Loandra and TT-OWI (all of them are downloaded from MSE2021). We apply the scoring function used in MSE2021 to evaluate the performance of these four solvers, since the scoring function is suitable for evaluating and comparing multiple solvers. The scoring function actually indicates how close the solutions are to the best-known solutions. Specifically, suppose $C_{BKS}$ is the cost of the best-known solution of an instance which is recorded in MSEs, $C_i$ is the cost of the solution found by the $i$-th solver ($i \in \{1,2,3,4\}$) in our experiments, the score of solver $i$ for this instance is $\frac{\mathop{min}(C_{BKS},C_1,C_2,C_3,C_4) + 1}{C_i + 1} \in [0,1]$ (resp. 0) if the solution found by solver $i$ is feasible (resp. infeasible). Finally, the score of a solver for a benchmark is the average value of the scores for all the instances in the benchmark. 

The comparison results of these four solvers are shown in Table \ref{table-BandMS-c}. BandMaxSAT-c yields the highest score on all the benchmarks except WPMS\_2019, demonstrating the excellent performance of our proposed method. 


\subsection{Ablation Study}
Finally, we perform ablation studies to analyze the effect of each component in BandMaxSAT. We first compare BandMaxSAT with two variants to evaluate the sampling strategy in the bandit model. The first one is BandMaxSAT$_\text{sample-1}$ (Sample-1 in brief), a variant of BandMaxSAT that sets the parameter $ArmNum$ to 1, which actually replaces the whole bandit model in BandMaxSAT with the simple random strategy used in Dist~\cite{Cai2014}, CCEHC~\cite{Luo2017}, SATLike~\cite{Lei2018} and SATLike3.0~\cite{Cai2020}. The second one is BandMaxSAT$_\text{sample-all}$ (Sample-all in brief), a variant of BandMaxSAT that removes the sampling strategy in the bandit model, i.e., selecting the arm to be pulled by traversing all the available arms. The results are shown in Table \ref{table-BandMS-sample}.

The results in Table \ref{table-BandMS-sample} show that our bandit model significantly outperforms the random strategy that is widely used in recent (W)PMS local search algorithms, and can greatly improve the performance. Moreover, the sampling strategy used in our bandit model is effective and necessary.

We then compare BandMaxSAT with its another variant BandMaxSAT$_{\text{no-delay}}$, that sets the parameter $d$ to 1, to evaluate the delayed reward method. The results are shown in Table \ref{table-BandMS-delay}. The results indicate that the delayed reward method fits well with the problems, and the method can help BandMaxSAT evaluate the quality of the arms better.

We further do two groups of experiments to evaluate the proposed HyDeci initialization method. The first group compares BandMaxSAT with its variant BandMaxSAT$_{\text{no-binary}}$, that does not prioritize binary clauses in HyDeci (i.e., remove lines 8-15 in Algorithm \ref{alg:HyDeci}). The second group compares two variants of BandMaxSAT. They are, BandMaxSAT$_{\text{fast}}$, a variant of BandMaxSAT that outputs the first feasible solution it found (within a time limit of 300 seconds), and BandMaxSAT$_{\text{no-binary-fast}}$, a variant of BandMaxSAT$_{\text{no-binary}}$ that outputs the first feasible solution it found. We actually use BandMaxSAT$_{\text{fast}}$ and BandMaxSAT$_{\text{no-binary-fast}}$ to roughly evaluate the quality of the initial assignments. The results of these two groups are shown in Tables \ref{table-BandMS-binary} and \ref{table-BandMS-fast}, respectively.

From the results in Tables \ref{table-BandMS-binary} and \ref{table-BandMS-fast} we can see that:

(1) BandMaxSAT$_{\text{fast}}$ outperforms BandMaxSAT$_{\text{no-binary-fast}}$ on all the benchmarks except WPMS\_2019, indicating that our method that prioritizes both unit and binary clauses can yield better initial assignments than the method that only prioritizes unit clauses. 

(2) BandMaxSAT outperforms BandMaxSAT$_{\text{no-binary}}$ for WPMS, indicating that our HyDeci method is effective and can improve the BandMaxSAT for WPMS. For PMS, These two algorithms have close performance, indicating that the local search process in BandMaxSAT is robust for PMS, as BandMaxSAT$_{\text{no-binary}}$ can obtain PMS solutions with similar quality to BandMaxSAT, with worse initial assignments.

\section{Conclusion}

This paper proposes a multi-armed bandit local search solver called \name for Partial MaxSAT (PMS) and Weighted PMS (WPMS). The proposed bandit model can help the local search learn to select an appropriate soft clause to be satisfied in the current step when the algorithm falls into a feasible local optimum. We further apply the sampling strategy and the delayed reward method to improve our bandit model. As a result, the bandit model fits well with (W)PMS. 
Moreover, we propose an effective initialization method, called HyDeci, that prioritizes both unit and binary clauses when generating the initial assignments. HyDeci can improve \name by providing high-quality initial assignments, and could be useful to improve other  local search MaxSAT solvers. 
Extensive experiments on all the (W)PMS instances from the incomplete tracks of the last four MSEs demonstrate that \name significantly outperforms the state-of-the-art local search (W)PMS algorithm SATLike3.0. Moreover, by combining \name with the complete solver TT-Open-WBO-Inc, the resulting solver \name-c outperforms the
complete solvers SATLike-c, Loandra, and TT-Open-WBO-Inc. 

The key issue in designing a local search algorithm is how to escape from local optima to find new high-quality search directions. In future work, we plan to generalize our bandit model to improve other local search algorithms for various NP-hard problems that need to select one among multiple candidates to escape from local optima.


\section*{Acknowledgements}
This work is supported by National Natural Science Foundation of China (62076105) and Grant PID2019-111544GB-C21 funded by MCIN/AEI/10.13039/501100011033.

\bibliographystyle{named}
\bibliography{ijcai22}

\end{document}